\documentclass[utf8]{frontiersSCNS} 

\usepackage[]{algorithm2e}
\usepackage{amsmath,amssymb,amsfonts}
\usepackage{booktabs}
\usepackage{cite}
\usepackage{graphicx}
\usepackage[hidelinks]{hyperref}
\usepackage{lineno}
\usepackage{microtype}
\usepackage{pbox}
\usepackage{subcaption}
\usepackage{textcomp}
\usepackage{url}
\usepackage{wrapfig}

\usepackage[onehalfspacing]{setspace}

\def\firstAuthorLast{Pape {et~al.}} 
\def\Authors{Constantin Pape\,$^{1,2}$, Alex Matskevych\,$^{1}$, Adrian Wolny\,$^{1,2}$, Julian Hennies\,$^{1}$, Giulia Mizzon\,$^{1}$, Marion Louveaux\,$^{3}$, Jacob Musser\,$^{1}$, Alexis Maizel\,$^{3}$, Detlev Arendt\,$^{1}$ and Anna Kreshuk\,$^{1,*}$}
\def\Address{$^{1}$EMBL Heidelberg \\
$^{2}$HCI University of Heidelberg \\
$^{3}$COS University of Heidelberg}

\def\corrAuthor{Anna Kreshuk}
\def\corrEmail{anna.kreshuk@embl.de}

\renewcommand{\thesubfigure}{\arabic{subfigure}}

\begin{document}
\onecolumn
\firstpage{1}

\title[Leveraging Domain Knowledge with Lifted Multicuts]{Leveraging Domain Knowledge to Improve Microscopy Image Segmentation with Lifted Multicuts} 

\author[\firstAuthorLast ]{\Authors} 
\address{} 
\correspondance{} 

\extraAuth{}

\maketitle

\begin{abstract}
The throughput of electron microscopes has increased significantly in recent years,
enabling detailed analysis of cell morphology and ultrastructure in fairly large tissue volumes. Analysis of neural circuits at single-synapse resolution remains the flagship target of this technique, but applications to cell and developmental biology are also starting to emerge at scale. On the light microscopy side, continuous development of light-sheet microscopes has led to a rapid increase in imaged volume dimensions, making Terabyte-scale acquisitions routine in the field.

The amount of data acquired in such studies makes manual instance segmentation, a fundamental
step in many analysis pipelines, impossible. 
While automatic segmentation approaches have improved significantly thanks to the
adoption of convolutional neural networks, their accuracy still lags behind human annotations
and requires additional manual proof-reading.
A major hindrance to further improvements is the limited field of view of the segmentation networks
preventing them from learning to exploit the expected cell morphology or other prior biological knowledge which
humans use to inform their segmentation decisions.
In this contribution, we show how such domain-specific information can be leveraged by expressing it as long-range interactions
in a graph partitioning problem known as the lifted multicut problem.
Using this formulation, we demonstrate significant improvement in segmentation accuracy for four challenging boundary-based
segmentation problems from neuroscience and developmental biology.
\end{abstract}

\section{Introduction}
Large-scale electron microscopy (EM) imaging is becoming an increasingly important tool in
different fields of biology. The technique was pioneered by the efforts to trace the neural circuitry 
of small animals at synaptic resolution to obtain their so-called connectome -- a map of neurons and synapses between them. In the 1980's White et al.\ \citep{white1986structure} have mapped the complete connectome of \textit{C. elegans} in a manual
tracing effort which spanned over a decade. Since then, the throughput of EM imaging has increased by several orders of magnitude thanks to innovations like multi-beam serial section EM \citep{eberle2015high}, hot-knife stitching \citep{hayworth2015ultrastructurally} or gas cluster milling \citep{hayworth2018serial}. 
This allows to image much larger volumes up to the complete brain of the fruit-fly larva \citep{eichler2017complete} and even the adult fruit-fly \citep{zheng2018complete}.
Recently, studies based on large-scale EM have become more common in other fields of biology as well \citep{nixon2016increased, otsuka2018postmitotic, russell20173d}.

In light microscopy, very large image volumes became routine even earlier \citep{royerAdaptive2016,krzicMultiview2012,Keller1065}, with Terabyte-scale acquisitions not uncommon for a single experiment. While the question of segmenting cell nuclei at such scale with high accuracy has been addressed before \citep{amatFast2014}, cell segmentation based on membrane staining remains a challenge and a bottleneck in analysis pipelines. 

Given the enormous amount of data generated, automated analysis of 
the acquired images is crucial; one of the key steps being instance segmentation of cells or cellular organelles.
In recent years, the accuracy of automated segmentation algorithms has increased significantly thanks to the rise of deep learning-based tools in computer vision and the development of convolutional neural networks (CNNs) for semantic and instance segmentation \citep{beier2017multicut,funke2018large,lee2017superhuman,januszewski2018high}. 
Still, it is not yet good enough to completely forego human proof-reading. Out of all microscopy image analysis problems, neuron segmentation in volume EM turned out to be particularly difficult \citep{januszewski2018high} due to the small diameter and long reach of neurons and astrocytes, but other EM segmentation problems have not yet been fully automated either. Heavy metal staining used in the EM sample preparation affects all cells indiscriminately and forces segmentation algorithms to rely on membrane detection to separate the objects. The same problem arises in the analysis of light microscopy volumes with membrane staining, where methods originally developed for EM segmentation also achieve state-of-the-art results \citep{Funke_2018_ECCV_Workshops}. 

One of the major downsides of CNN-based segmentation approaches lies in their limited field of view which makes them overly reliant on local boundary evidence. Staining artifacts, alignment issues or noise can severely weaken this evidence and often cause \textit{false merge errors} where separate objects get merged into one. 
On the other hand, membranes of cellular organelles or objects of small diameter often cause \textit{false split errors} where a single structure gets split into several segmented objects. 

Human experts avoid many of these errors by exploiting additional prior knowledge on the expected object shape or constraints from higher-level biology. Following this observation, several algorithms have recently been introduced to enable detection of morphological errors in segmented objects \citep{rolnick2017morphological,zung2017error,dmitriev2018efficient,matejek2019biologically}. By looking at complete objects rather than a handful of pixels, these algorithms can significantly improve the accuracy of the initial segmentation. In addition to purely morphological criteria, Krasowski et al.\ in \citep{krasowski2017neuron} suggested an algorithm to exploit biological priors such as an incompatible mix of ultrastructure elements.

Building on such prior work, this contribution introduces a general approach to leverage domain-specific knowledge for the improvement of segmentation accuracy. It allows to incorporate a large variety of rules, explicit or learned from data, which can be expressed as the likelihood of certain areas in the image to belong to the same object in the segmentation. The areas can be \emph{sparse} and/or \emph{spatially distant}. In more detail, we formulate the segmentation problem with such rules as a graph partitioning problem with long-range attractive or repulsive edges.

\begin{figure}[ht]
    \centering
    \includegraphics[width=0.75\textwidth]{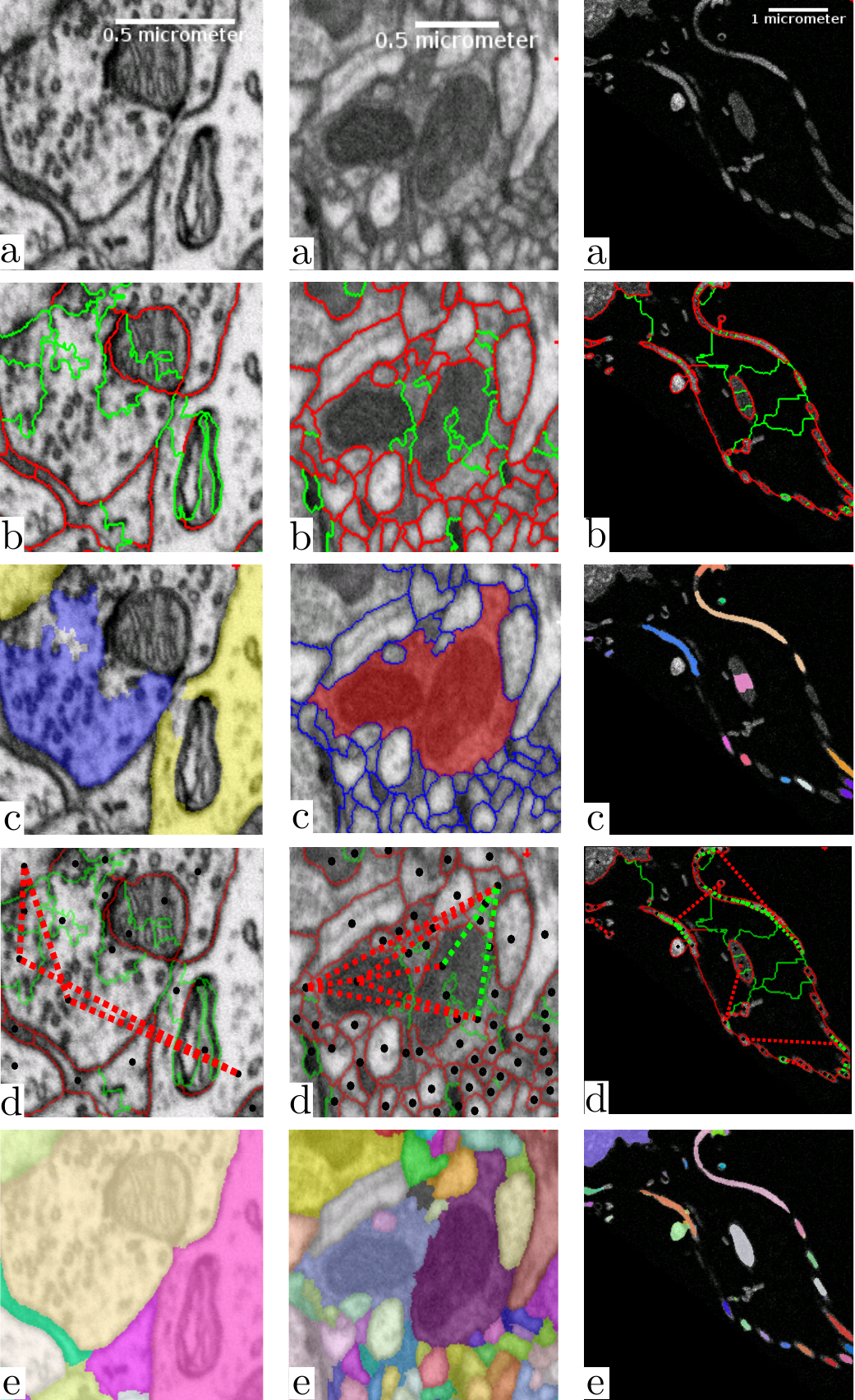}
\end{figure}
\clearpage
    \captionof{figure}{Mapping domain knowledge to sparse lifted edges for mammalian cortex (left), drosophila brain (middle) and sponge choanocyte chamber (right). The raw data is shown in (a). Based on local boundary evidence (not shown) predicted by a Random Forest or a CNN, we group the volume pixels into superpixels, which form a region adjacency graph. The edges of the graph correspond to boundaries between the superpixels as shown in (b). The edges are weighted, with weights derived from boundary evidence or predicted by an additional classifier (see also \ref{tab_res_overview}). Weights can make edges attractive (green) or repulsive (red). (c) shows the domain knowledge mapped to superpixels: axon (blue) and dendrite (yellow) attributions (left); an object with implausible morphology (red, center); semantically different objects (one color per object, right). Superpixels with mapped domain knowledge are connected with lifted edges as shown in (d), with green for attractive and red for repulsive edges (only a subset of edges is shown to avoid clutter). (e) displays the solution of the complete optimization problem with local and sparse lifted edges as the final segmentation (e). }
    \label{methods_figure}

\hrule\bigskip
For the problem of image segmentation, the graph in the partitioning problem corresponds to the region adjacency graph of the image pixels or superpixels. The nodes of the graph can be mapped directly to spatial locations in the image. When domain knowledge can be expressed as rules that certain locations must or must not belong to the same object, it can be distilled into lifted (long-range) edges between these locations. The weights of such lifted edges are derived from the strictness of the rules, which can colloquially range from "usually do / do not belong to the same object" to "always / never ever belong to the same object". 

Horvnakova et al.\ in \citep{horvnakova2017analysis} showed that this problem, which they term Lifted Multicut as it corresponds to the Multicut partitioning problem with additional edges between non-adjacent nodes, can be solved exactly in reasonable time for small problem sizes, while Beier et al.\ in \citep{beier2016efficient} introduced an efficient approximate solver.

In the following, we demonstrate the versatility of the Lifted Multicut based-approach by applying it to four segmentation problems, three in EM and one in LM. We incorporate starkly different kinds of prior information into this framework:
\begin{itemize}
\item Based on the knowledge that axons are separated from dendrites in mammalian cortex, we use indicators of axon/dendrite attribution to avoid merges between axonal and dendritic neural processes (\autoref{methods_figure}(left));

\item Based on the knowledge of plausible neuron morphology, we correct false merge errors in the segmentation of neural processes (\autoref{methods_figure}(center));

\item Based on the knowledge that certain biological structures form long continuous objects, we reduce the number of false splits in instance segmentation of sponge choanocytes (\autoref{methods_figure}(right));

\item Based on the knowledge that a cell should only contain one nucleus, we improve the segmentation of growing plant lateral roots (\autoref{plant_figure}).
\end{itemize}

Aiming to apply the method to data of biologically relevant size, we additionally introduce a new scalable solver for the lifted multicut problem based on our prior work from \citep{pape2017solving}. 
Our code is available at \url{https://github.com/constantinpape/cluster_tools}.

\section{Related Work} \label{sec_related_work}


Neuron segmentation for connectomics has been the main driver of the recent advances in boundary-based segmentation for microscopy. Most methods (\citep{andres2012globally, beier2017multicut, nunez2013machine, funke2018large, lee2017superhuman}) follow a three step procedure: 
in the first step they segment boundaries, in the second compute an over-segmentation into superpixels and finally agglomerate the superpixels into objects.

The success of a CNN \citep{ciresan2012deep} in the ISBI 2012 neuron segmentation challenge \citep{arganda2015crowdsourcing} has prompted the adoption of this technique for the boundary prediction step.
Most recent approaches use a U-Net \citep{ronneberger2015u} architecture and custom loss functions \citep{lee2017superhuman, funke2018large}. 
The remaining differences between methods can be found in the superpixel merging procedure. Several approaches are based on hierarchical clustering, but differ in how they accumulate boundary weights: \citep{lee2017superhuman} use the accumulated
mean CNN boundary predictions,
\citep{funke2018large} employ quantile based accumulation and \citep{nunez2013machine} re-predict the weights with a random forest \citep{breiman2001random} after each agglomeration step.
In contrast \citep{andres2012globally} and \citep{beier2017multicut} solve a NP-hard graph partitioning problem, the (Lifted) Multicut.
Notable exception from this three step approach are the flood filling network (FFN) \citep{januszewski2018high} and MaskExtend \citep{meirovitch2016multi} which can go directly from pixels to instances by predicting individual object masks one at a time, as well as 3C \citep{meirovitch2018cross}, which can simultaneously predict multiple objects. 


Krasowski et al \citep{krasowski2017neuron} showed that the common three-step procedure can be modified to incorporate sparse biological priors at the superpixel agglomeration step. They use the Asymmetric Multi-Way Cut (AMWC) \citep{kroeger2014asymmetric}, a generalization of the Multicut for joint graph partition and node labeling. The method is based on exploiting the knowledge that, given the field of view of modern electron microscopes, axon- and dendrite-specific ultrastructure should not belong to the same segmented objects in mammalian cortex. While this approach can be generalized to other domain knowledge, it has two important drawbacks. First, it is not possible to encode attractive information just with node labels.
Second, it is harder to express information that does not fit the node labeling category, even if it is repulsive in nature. A good example for this is the 
morphology-based false merge correction. In this case, defining a labeling for only a subset of nodes is not possible.

Lifted Multicut formulation has been used for neuron segmentation by \citep{beier2017multicut}. However, the lifted edges were added densely and their weights and positions were not based on domain knowledge, but learned from groundtruth segmentations by the Random Forest algorithm. Only edges over a graph distance of 3 were considered. 
These lifted edges made the segmentation algorithm more robust against single missing boundaries, but did not counter the problem of the limited field of view of the boundary CNN and did not prevent biologically implausible objects. Note that this approach can be seen as a special case of the framework proposed here, using generic, but weak knowledge about
local morphology and graph structure of segments.
Besides Lifted Multicut, the recently introduced Mutex Watershed \citep{wolf2018mutex, wolf2019mutex} and generalized agglomerative clustering \citep{Bailoni2019AGF} can also exploit long-range information. 

While all the listed methods demonstrate increased segmentation accuracy, they do not offer a general recipe on how to exploit domain-specific knowledge in a segmentation algorithm. We propose a versatile framework that can incorporate such information from diverse sources by mapping it to sparse lifted edges in the lifted multicut problem.

\section{Methods} \label{sec4}

Our method follows the three step segmentation approach described in \autoref{sec_related_work}, starting from a boundary predictor and using graph partitioning to agglomerate super-pixels.
First, we review the lifted multicut problem \citep{horvnakova2017analysis} in \autoref{methods:graph}. We follow by proposing a general approach to incorporate domain-specific knowledge into the lifted edges (\autoref{methods:sparse_edges}). Finally, we describe four specific applications with different sources of domain knowledge and show how our previous work on lifted multicut for neuron segmentation can be positioned in terms of the proposed framework.

\subsection{Lifted Multicut Graph Partition} \label{methods:graph}

Instance segmentation can be formulated as a graph partition problem given a graph $G = \{V, E\}$ and edge weights $W \in ]-\infty, \infty[$ . 
In our setting, the nodes $V$ correspond to fragments of the over-segmentation and edges $E$ link two nodes iff the two corresponding fragments share an image boundary.
The weights $W$ encode the attractive strength (positive values) or repulsive strength (negative values) of edges
and are usually derived from (pseudo) probabilities $P$ via negative log-likelihood:
\begin{equation} \label{eq_log}
w_e = \textrm{log} \frac{1 - p_e}{p_e} ~ \forall ~ e \in E.    
\end{equation}
The resulting partition problem is known as multicut or correlation clustering \citep{chopra1993partition,andres2012globally,demaine2006correlation}. Its objective is given by
\begin{align} \label{eq_mc}
    & \min_{y_e \in Y_{E}} \sum_{e \in E} w_e y_e \quad \textrm{under the constraints} \\
    & \forall C \in \textrm{cycles}(G) \forall e \in C: y_e \leq \sum_{\hat e \in C \setminus \{e\}} y_{\hat e},
\end{align}
where $Y_E$ are binary indicator variables linked to the edge state; 0 means that an edge connects the two adjacent nodes in the resulting partition, 1 means that it doesn't.
The constraints forbid \textit{dangling edges} in the partition, i.e. edges that separate two nodes ($y_e = 1$) for which a path of connecting edges ($y_e = 0$) exists.

The lifted multicut \citep{horvnakova2017analysis} is an extension of the multicut problem, which introduces a new set of edges $F$ called lifted edges.
These edges differ from regular graph edges by providing only an energy contribution, but not inducing connectivity.
This is motivated by the observation that it is often helpful to derive non-local features for the connectivity of (super) pixels.
The presence of an attractive non-local edge should not result in \textit{air bridges} though, i.e. non-local edges that connect two pixels without a connection via local edges.
In our setting, lifted edges connect nodes $v$ and $w$ that are not adjacent in $G$.
With the sets of original edges $E$, lifted edges $F$, binary indicator variables $Y$, and weights $W$ associated with all edges in ${E \cup F}$ the lifted multicut objective can be formulated as

\begin{align} \label{eq_lmc}
    & \min_{y_e \in Y_{EF}} \sum_{e \in E \cup F} w_e y_e \quad \textrm{under the constraints} \\
    & \forall C \in \textrm{cycles}(G) \forall e \in C: y_e \leq \sum_{\hat e \in C \setminus \{e\}} y_{\hat e} \\
    & \forall vw \in F \forall P \in vw-\textrm{paths}(G): y_{vw} \leq \sum_{e \in P} y_e \\
    & \forall vw \in F \forall c \in vw-\textrm{cuts}(G):1 - y_{vw} \leq \sum_{e \in C} (1 - y_e).
\end{align}

The constraints (5) correspond to \autoref{eq_mc} and enforce a consistent partition without dangling edges.
Constraints (6) and (7) ensure that the state of lifted edges
is consistent with the connectivity, i.e. that two nodes connected by a lifted edge
are also connected via a path of regular edges and two nodes separated by a lifted edge are not connected
by any such path.

\begin{figure}[ht]
    \centering
    \includegraphics[width=.24\textwidth]{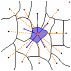}
    \includegraphics[width=.24\textwidth]{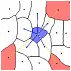}
    \caption{(Left) Graph neighborhood of a single node (blue shaded segment) with local edges (blue lines) and dense lifted edges (orange dotted edges). (Right) Neighborhood with sparse lifted edges (yellow dotted edges), connecting nodes with projected domain knowledge (red shaded segments).}
    \label{fig_toy_graph}
\end{figure}

\subsection{Sparse Lifted Edges} \label{methods:sparse_edges}

Our main contribution is a general recipe how to express domain-specific knowledge via sparse lifted edges
that are only added between graph nodes where attribution of this knowledge is possible.
The right part of \autoref{fig_toy_graph} shows a sketch of this idea: nodes with attribution are shown by shaded segments and sparse lifted edges by yellow dashed lines.  

The sparse lifted edges are constructed in several steps, see \autoref{methods_figure}. We compute the superpixels by running the watershed algorithm on boundary predictions and construct the corresponding region adjacency graph. \autoref{methods_figure}(b) shows regular, not lifted, edges between superpixels, green for attractive and red for repulsive weights. 
Then, we map the domain specific knowledge to nodes as shown in \autoref{methods_figure}(c), and derive attractive and repulsive lifted edges, again shown as green and red lines in (d).
The sign and strength of the lifted edge weights can be learned or introduced explicitly, reflecting the likelihood of incident nodes being connected. \autoref{eq_log} is used to obtain signed weights.
Finally, we  solve the resulting lifted multicut objective to obtain an instance segmentation, shown in \autoref{methods_figure}(e). 



\paragraph{\textbf{Mouse Cortex Segmentation, EM}}
This application example shows how the framework described above can be used to incorporate the axon/dendrite attribution priors first introduced in \citep{krasowski2017neuron}. We detect the axon- and dendrite-specific elements and map them to the nodes in the same way as \citep{krasowski2017neuron} (\autoref{methods_figure}(c), with blue shading for axon and yellow for dendrite attribution). The difference comes in the next step: instead of introducing semantic node labels for "axon" and "dendrite" classes, we add repulsive lifted edges between nodes which got mapped differently. \autoref{res_axondendrite} includes more details on the problem set-up and results.


\paragraph{\textbf{Drosophila brain segmentation, EM}}
For neurons in the insect brain, the axon/dendrite separation is not pronounced and the approach described in the previous section can not be applied directly.
Instead, morphological information can be used to identify and resolve errors in segmented objects. This was first demonstrated by \citep{rolnick2017morphological}, where a CNN was trained on downsampled segmentation masks to detect merge errors. Meirovitch et al.\ in \citep{meirovitch2016multi} detect merge errors with a simple shape-based heuristic and then correct these with a MaskExtend algorithm. Zung et al. \citep{zung2017error} were the first to combine CNN-based error detection and flood filling network-based correction. In their formulation both false merges and false splits can be corrected. Recently \citep{dmitriev2018efficient, matejek2019biologically} have introduced an approach based on CNN error detection followed by a simple heuristic to correct false merges and lifted multicut graph partitioning to fix false splits.

Based on all this prior work which convincingly demonstrates that false merge errors can be detected in a post-processing step, we concentrate our efforts on error \emph{correction}, emulating the detection step with an oracle. We extract skeletons for all segmented objects and have the oracle predict, for all paths connecting terminal nodes of a skeleton, if this path goes through a false merge location (passes through an unidentified boundary). Note that the oracle is not perfect and we evaluate the performance of the algorithm for different levels of oracle error. 

If the oracle predicts the path to go through a false merge, we introduce a repulsive lifted edge between the terminals of the path. The weights of the edges are also predicted by the oracle. \autoref{methods_figure} shows an example of this approach: the red object in the middle of panel (c) has been detected as a false merge. The corresponding lifted edges are shown in \autoref{methods_figure}(d). 


\paragraph{\textbf{Sponge segmentation, EM}}
In this example, we tackle a segmentation problem in a sponge choanocyte chamber. These structures are built from several surrounding cells, the choanocytes,
that interact with a central cell via flagella which are surrounded by a collar of microvilli. Our goal is to segment cell bodies, flagella and microvilli.
This task is challenging due to the large difference in sizes of these structures.
Especially the segmentation of the small flagella and microvilli is difficult. Without the use of domain specific knowledge on their continuity, the Multicut algorithm splits them up into many pieces.

In order to alleviate these false split errors, we predict which pixels in the image belong to flagella and microvilli and compute an approximate flagella and microvilli instance segmentation via thresholding and connected components.
We map the component labels to nodes of the graph, see right column in \autoref{methods_figure}(c). Then, we introduce attractive lifted edges between
the nodes that were covered by the same component and repulsive lifted edges between nodes mapped to different components, see \autoref{methods_figure}(d).

\paragraph{\textbf{Lateral root segmentation, LM}}
Finally, we tackle a challenging segmentation problem in light-sheet data: segmentation of root cells in \textit{Arabidopsis thaliana}.
This data was imaged with two channels, showing cell membrane and nucleus markers. We use the first channel to predict cell boundaries and the second to segment individual nuclei. The nuclei then serve as bases to force each segmented cell to only contain one nucleus: we introduce repulsive lifted edges between nodes which are covered by different nuclei instances. \autoref{res_plants} shows how this setup helps prevent false merge errors in cell segmentation.


\subsection{Hierarchical Lifted Multicut Solver} \label{sec_solver}

Finding the optimal solution of the lifted multicut objective is NP-hard. Approximate solvers based on greedy algorithms \citep{keuper2015efficient} and fusion moves \citep{beier2017multicut} have
been introduced. However, even these approximations do not scale to the large problem we need to solve in the sponge segmentation example. 
In order to tackle this and even larger problems, we adapt the hierarchical multicut solver of \citep{pape2017solving} for lifted multicuts.

This solver extracts sub-problems from a regular tiling of the volume, solves these sub-problems in parallel and uses the solutions to contract nodes in 
the graph, thus reducing its size. This approach can be repeated for an increasing size of the blocks that are used to tile the volume,
until the reduced problem becomes feasible with an other (approximate) solver.

We extend this approach to the lifted multicut by also extracting lifted edges during the sub-problem extraction. We only extract
lifted edges that connect nodes in the sub-graph defined by the block at hand. This strategy, where we ignore lifted edges
crossing block boundaries, is in line with the idea that lifted edges contribute to the energy, but not to the connectivity.
Note that lifted edges that are not part of any sub-problem at a given level will still be considered at a later stage.
See appendix \autoref{algo_lmc} for pseudo-code. The comparison to other solvers in appendix \autoref{tab_res_lmc} shows that 
it indeed scales better to large data.
Note that this approach is conceptually similar to the fusion move based approximation of \citep{beier2016efficient}, which extracts and solves
sub-problems based on a random graph partition and accepts changes from sub-solutions if they increase the overall energy, repeating this process until convergence.
Compared to this approach, we extract sub-problems from a deterministic partition of the graph. This allows us to solve only a preset number of sub-problems leading
to faster convergence. 

Note that our approximate solver is only applicable if the graph at hand has a spatial embedding, which allows to extract sub-problems from a tiling of space. 
In our case, this spatial embedding is given by the watershed fragments that correspond to nodes.

\section{Results} \label{sec5}

We study the performance of the proposed method on four different problems: i) neuron segmentation in murine cortex with priors from axon/dendrite segmentation, ii) neuron segmentation in drosophila brain with priors from morphology-based error detection, iii) instance segmentation in a sponge choanocyte chamber with priors from semantic classes of segmented objects, ix) cell segmentation in Arabidopsis roots with priors from "one nucleus per cell" rule.
Appendix \autoref{tab_res_overview} summarizes the different problem set-ups.
We evaluate segmentation quality using the variation of information (VI) \citep{meilua2003comparing}, which can be separated into split and merge scores, and the adapted rand score \citep{arganda2015crowdsourcing}.
For all error measures used here, a lower value corresponds to higher segmentation quality.

\subsection{Mouse Cortex Segmentation, EM} \label{res_axondendrite}

We present results on a volume of murine somatosensory cortex that was acquired by FIBSEM at 5 $\times$ 5 $\times$ 6 nanometer resolution. 
The same volume has already been used in \citep{krasowski2017neuron} for a similar experiment. To ensure a fair comparison between the two methods for incorporating axon/dendrite priors, we obtained derived data
from the authors and use it to set-up the segmentation problem.

This derived data includes probability maps for cell membrane, mitochondria, axon and dendrite attribution
as well as a watershed over-segmentation derived from the cell membrane probabilities and ground-truth instance segmentation.
From this data, we set up the graph partition problem as follows: we build the region adjacency graph $G$ from the watersheds and 
compute weights for the regular edges with a random forest based on edge and region appearance features. See \citep{beier2017multicut} 
for a detailed description of the feature set. Next, we introduce dense lifted edges up to a graph distance of three. We use a random forest
based on features derived from region appearance and clustering to predict their weights, see \citep{beier2017multicut} for details.
In addition to the region appearance features only based on raw data, we also take into account the mitochondria attribution here.
Next, we map the axon/dendrite attribution to the nodes of $G$ and introduce sparse lifted edges between nodes mapped to \emph{different} classes.
We infer weights for these edges with a random forest based on features from the statistics of the axon and dendrite node mapping.
We use the fusion move solver of \citep{beier2016efficient} for optimizing the lifted multicut objective.

We divide the volume into a 1 $\times$ 3.5 $\times$ 3.5 micron block that is used to train the random forests for edge weights
and a 2.5 $\times$ 3.5 $\times$ 3.5 micron block used for evaluation.
The random forest predicting pixel-wise probabilities was trained by the authors of \citep{krasowski2017neuron}
on a separate volume, using ilastik \citep{sommer2011ilastik}.

We compare the multicut and AMWC solutions reported in \citep{krasowski2017neuron} with different
variants of our methods, see \autoref{tab5_1}.
As a baseline, we compute the lifted multicut only with dense lifted edges and without features
from mitochondria predictions (LMC-D).
We compute the full model with dense and sparse lifteds (LMC-S) with and without additional features for dense lifted edges from mitochondria predictions.
In addition, we compare to an iterative approach (LMC-SI) similar to the error correction approach in \autoref{res_falsmerge}, where we perform LMC-D segmentation first and introduce sparse lifted edges 
only for objects that contain a false merge (identified by presence of both axonic and dendritic nodes in the same object).

The LMC-D segmentation quality is on par with the AMWC, although it does not use any 
input from the priors, showing the importance of dense lifted edges. 
Our full model with sparse lifted edges shows significantly better quality compared to LMC-D.
Mitochondria-based features provide a small additional boost. The segmentation quality of the iterative approach
LMC-SI is inferior to solving the full model LMC-S. This shows the importance of joint optimization of the full model with dense and sparse
lifted edges.

\begin{table}[h]
    \center
    \begin{tabular}{l r r r}
        \toprule
        Method                         & VI-Split  & VI-Merge  & Rand Error \\
		\midrule
        MC \citep{krasowski2017neuron}  & \textbf{0.3471}    & 0.6347    & 0.0787     \\
        AMWC \citep{krasowski2017neuron}& 0.4578    & 0.4935    & 0.0754     \\ 
        LMC-D                          & 0.4144    & 0.4445    & 0.0891     \\
        LMC-S                          & 0.4133    & \textbf{0.3788}    & \textbf{0.0362}     \\
        LMC-S (No Mitos)               & 0.4038    & 0.3966    & 0.0363     \\
        LMC-SI                         & 0.5054    & 0.3998    & 0.0586     \\
		\bottomrule
    \end{tabular}
    \caption{Variants of our approach compared to the method of \citep{krasowski2017neuron}. The Rand Error measures the over-all segmentation quality, while VI-Split measures the degree of over-segmentation and VI-Merge the degree of under-segmentation. For all measures, a lower score corresponds to a better segmentation.}
    \label{tab5_1}
\end{table}

\subsection{Drosophila brain segmentation, EM} \label{res_falsmerge}
We test the false merge correction on parts of the \textit{Drosophila} medulla using a 68 $\times$ 38 $\times$ 44 micron FIBSEM volume imaged at 8 $\times$ 8 $\times$ 8 nanometer from \citep{takemura2015synaptic}, who also provide a ground-truth segmentation for the whole volume.

First, we train a 3D U-Net for boundary prediction on a separate 2 $\times$ 2 $\times$ 2 micron cube. 
We use this network to predict boundaries on the whole volume, and run watershed over-segmentation based on these predictions.
Then, we set up an initial Multicut with edge weights derived from mean accumulated boundary evidence.
We obtain an initial segmentation by solving it with the block-wise solver of \citep{pape2017solving}.

In order to demonstrate segmentation improvement based on morphological features, we skeletonize all sufficiently large objects using the method of \citep{lee1994building} implemented in \citep{van2014scikit}.
We then predict false merges along all paths between skeleton terminal nodes, using the groundt-ruth segmentation as oracle predictor.
Note that \citep{dmitriev2018efficient} have shown that it is possible to train a very accurate CNN to classify false merges based on morphology information in this set-up.
Given these predictions, we set up the Lifted Multicut problem by selecting all objects that have at least one path with a false merge detection.
For these objects, we introduce lifted edges between \emph{all} terminal nodes corresponding to paths and derive weights for these edges from the false merge 
probability (note that we use an imperfect oracle for some experiments, so the merge predictions are not absolutely certain).
We solve two different variants of this problem, LMC-S, where we solve the whole problem using the solver introduced in \autoref{sec_solver} and LMC-SI, where we only solve
the sub-problems arising for the individual objects.
For this, we use the Fusion Moves solver of \citep{beier2016efficient}.

\autoref{tab_res_fib} compares the results of the initial Multicut (MC) with LMC-S and LMC-SI (using a perfect oracle) as well as the current state of the art FFN based segmentation \citep{januszewski2018high}.
We adopt the evaluation procedure of \citep{januszewski2018high} and use a cutout of size 23 $\times$ 19 $\times$ 23 micron for validation. We use two different versions of the ground-truth, the full segmentation and only a set of white-listed objects that were more carefully proofread. The FFN segmentation and validation ground-truth was kindly provided by the authors of \citep{januszewski2018high}.
The results show that our initial segmentation is inferior to FFN in terms of merge errors, but using LMC-SI we can improve the merge error to be even better than FFN.
Interestingly, LMC-SI performs better than LMC-S. We suspect that this is due to the fact that we only add lifted edges inside of objects with a false merge detection, thus 
LMC-S does not see more information then LMC-SI, while having to solve a much bigger optimization problem.

\begin{figure}[h]
    \centering
    \includegraphics[width=.775\textwidth]{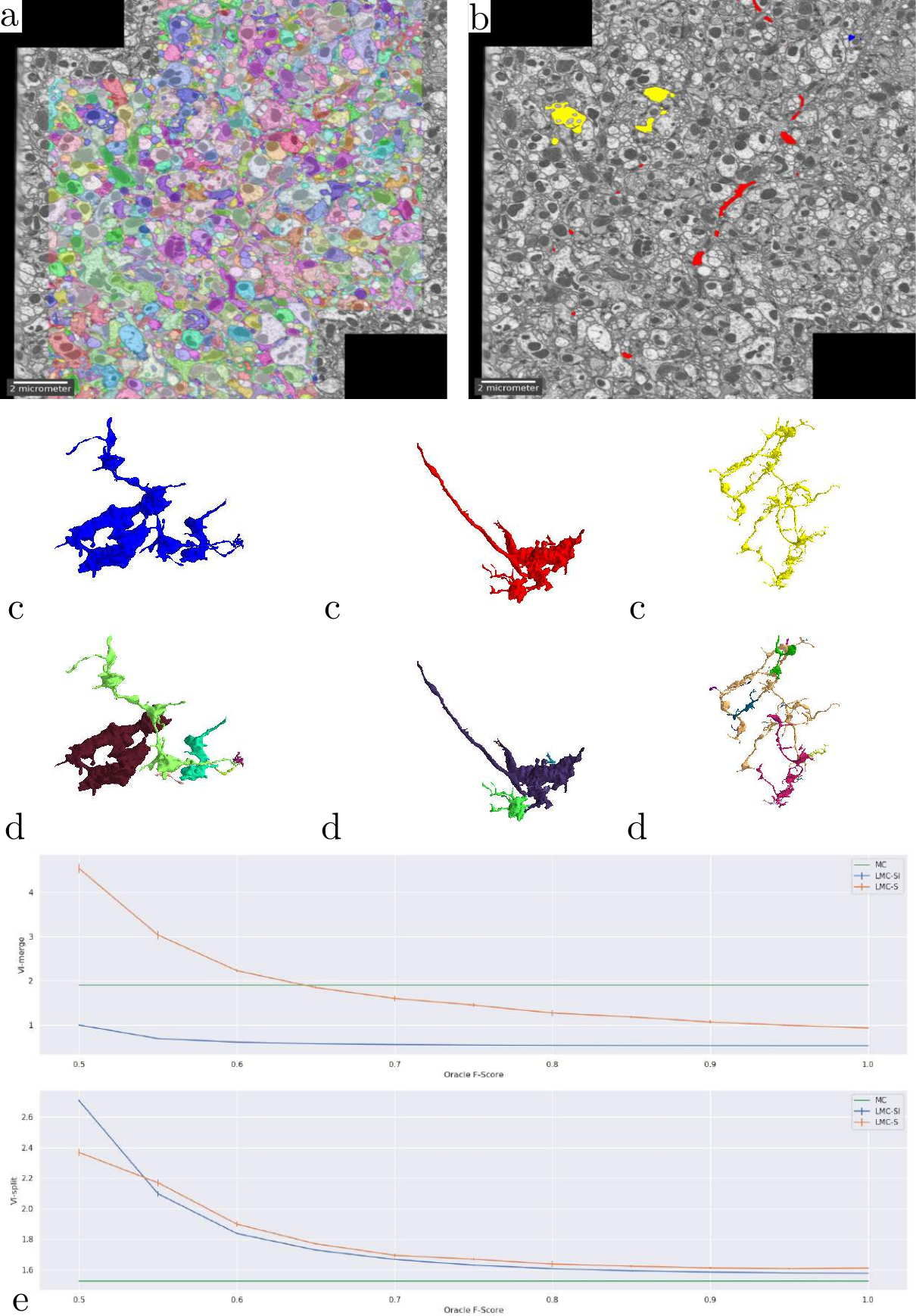}
    \caption{Overview of results on the drosophila medulla dataset. We detect merges in the initial segmentation result (a) using an oracle. The red, blue and yellow segments in (b) were flagged as false merges. (c) and (d)  show merged / correctly resolved objects. (e) shows the performance of our approach when tuning the F-Score of our oracle predictor from 0.5 to 1.}
    \label{fib_figure}
\end{figure}

In \autoref{fib_figure} we show the initial segmentation and three corrected merges. Panel (e) evaluates LMC-S and LMC-SI on the full ground-truth when using an imperfect oracle:
we tune the oracle's F-score from 0.5 to 1.0 and measure VI-split and VI-merge. The curves show that LMC-SI is fairly robust against noise in the oracle predictions; it starts with a lower VI-merge than the initial MC, even for F-Score 0.5 and its VI-split gets close to the MC value for F-Score 0.75+.  

\begin{table}[t]
    \center
    \begin{tabular}{l r r r r}
        \toprule
                      & Full      &           & Whitelist & \\
                      & VI-split  & VI-merge  & VI-split & VI-merge \\
        \midrule
        MC            & 1.5246    & 1.9057    & 1.2189   & 0.6532   \\
        LMC-S         & 1.6110    & 0.9405            & 1.3050   & 0.2544   \\
        LMC-SI        & 1.5773    & \textbf{0.5403}   & 1.2369          & \textbf{0.0122}   \\
        FFN           & \textbf{1.4653}    & 0.6340   & \textbf{0.8702} & 0.0559   \\  
        \bottomrule
    \end{tabular}
    \caption{Results on the drosophila medulla dataset. We compare the segmentation results of Multicut (MC), Lifted Multicut solved for the whole volume (LMC-SI) and Lifted Multicut solved separately for all sub-problems arising from falsely merged objects (LMC-SI) with the results of FFN form \citep{januszewski2018high}. We use a cutout for validation and evaluate with the complete ground-truth segmentation (Full) and a subset of closely proof-read objects (Whitelist).}
    \label{tab_res_fib}
\end{table}

\vspace{0.5cm}
 \subsection{Sponge segmentation, EM} \label{res_sponge}

The two previous experiments mostly profited from repulsive information derived from ultrastructure or morphology.
In order to show how attractive information can be exploited, we turn to an instance segmentation problem in a sponge choanocyte chamber.
The EM volume was imaged with FIBSEM at a resolution of 15 $\times$ 15 $\times$ 15 nanometer.
We aim to segment structures of three different types: cell bodies, flagella and microvilli.
Flagella and microvilli have a small diameter, which make them difficult to segment with a boundary based approach.
On the other hand, cell bodies have a much larger diameter and touch each other, which makes a boundary based approach appropriate.

In order to set-up the segmentation problem, we first compute probability maps for boundaries, microvilli and flagella attribution
using the autocontext workflow of ilastik \citep{sommer2011ilastik}.
We set-up the lifted multicut problem by first computing watersheds based on boundary maps, extracting the region adjacency graph
and computing regular edge weights from the boundary maps accumulated over the edge pixels. We do not introduce dense lifted 
edges. For sparse lifted edges, we compute an additional instance segmentation of flagella and microvilli by thresholding the corresponding
probability maps and running connected components. Then, we map the components of this segmentation to graph nodes and connect nodes mapped to the same component via attractive lifted edges and nodes mapped to different components via repulsive lifted edges.
We use the hierarchical lifted multicut solver introduced in \autoref{sec_solver} to solve the resulting objective, using the approximate solver of \citep{keuper2015efficient} to solve sub-problems.
Note that the full model contained too many variables to be optimized by any other solver in a reasonable amount of time.

We run our segmentation approach on the whole volume, which covers a volume of 70 $\times$ 75  $\times$ 50 microns, corresponding to 4600 $\times$ 5000 $\times$ 3300 voxels. 
For evaluation, we use three cutouts of size 15 $\times$ 15 $\times$ 1.5 microns with ground-truth for instance and semantic segmentation. We split the evaluation 
into separate scores for objects belonging to the three different structures, extracting them based on the semantic segmentation ground-truth.
See \autoref{tab_res_sponge} for the evaluation results, comparing the sparse lifted multicut (LMC) to the multicut baseline (MC).
As expected the quality of the segmentation of cell bodies is not affected, because we don't introduce lifted edges for those.
The split rate in flagella and microvilli decreases significantly leading to a better overall segmentation for these structures.

\begin{table}[t]
    \center
    \begin{tabular}{l r r r}
        \toprule
        Method     & VI-Split  & VI-merge  & Rand Error \\
        \midrule
        Cells &&&\\
        MC         & 0.6058    & 0.0116    & 0.0783     \\
        LMC        & 0.6004    & 0.0116    & 0.0782     \\
        \midrule
        Flagella &&&\\
        MC          & 0.4728    & 0.0812    &  0.1205    \\
        LMC         & 0.2855    & 0.0812    &  0.0429    \\
        \midrule
        Microvilli &&&\\
        MC         & 3.1760    & 1.1101    & 0.7409     \\
        LMC        & 2.2745    & 1.1807    & 0.6973     \\
        \bottomrule
    \end{tabular}
    \caption{Quality of the sponge chonanocyte segmentation for the three different types of structures.}
    \label{tab_res_sponge}
\end{table}

\vspace{0.5cm}
\subsection{Lateral root segmentation, LM} \label{res_plants}

We segment cells in light-sheet image volumes of the lateral root primordia of \textit{Arabidopsis thaliana}.
The time-lapse video consisting of 51 time points was obtained in vivo in close-to-natural growth conditions.
Each time point is a 3D volume of size 2048 $\times$ 1050 $\times$ 486 voxels each with resolution 0.1625 $\times$ 0.1625 $\times$ 0.25 micron.
The volume has two channels, one showing membrane marker, the other nucleus marker.
We work on two selected time points, namely: $T_{45}$ and $T_{49}$ taken from the later stages of development where the instance segmentation problem is more challenging due to growing number of cells.
The time points have dense ground-truth segmentation for a 1000 $\times$ 450 $\times$ 200 voxels cutout centered on the root primordia. Both cells and nuclei ground truth are available.

A variant of 3D U-Net \citep{CicekALBR16} was trained in order to predict cell membranes and nuclei respectively. The two networks were trained on dense ground-truth from time points which were not part of our evaluation.
Apart from the primary task of predicting membranes and nuclei respectively, both networks were trained on an auxiliary task of predicting long-range affinities similarly to \citep{lee2017superhuman} which proved to improve the effectiveness of the main task.

Using these networks, we predict cell boundary probabilities and nucleus foreground probabilities.
We use the nucleus predictions to obtain a nucleus instance segmentation by thresholding the probability maps at $p_{threshold}=0.9$ and running connected components analysis.

We compute superpixel from the watershed transform on the membrane predictions and compute weights for the regular edges via mean accumulated boundary evidence.
We set up lifted edges by mapping the nucleus instances to superpixels and connecting all nodes whose superpixels were mapped to different nuclei with repulsive lifted edges.

\begin{figure}[h]
    \centering
    \includegraphics[width=.775\textwidth]{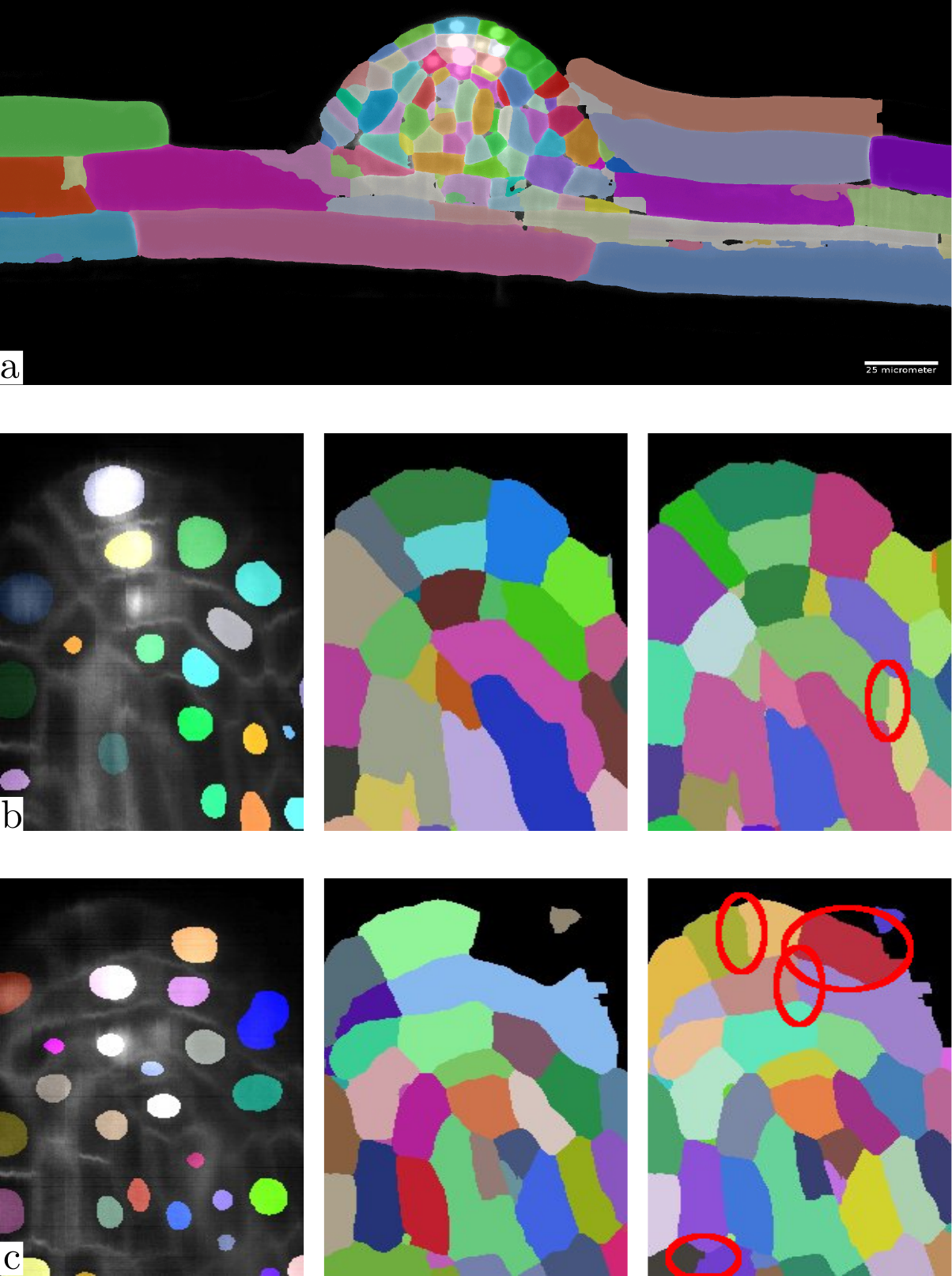}
    \caption{Overview of results on the plant root dataset. (a) shows one complete image plane with membrane channel and overview of the LMC segmentation for timepoint 49. (b) and (c) show zoom ins of the yz plane with raw data and nucleus segmentation (left), MC segmentation (middle) and LMC segmentation (right) with avoided merge errors marked.}
    \label{plant_figure}
\end{figure}

\autoref{tab_res_plant} shows the evaluation of segmentation results on the ground-truth cutouts. We can see that LMC-S clearly improves the merge errors as well the overall Rand Error while only marginally diminishing the split quality.
See \autoref{plant_figure} for an overview of the qualitative results.

\begin{table}
    \center
    \begin{tabular}{l r r r r r r}
        \toprule
                     & MC       &          &            & LMC-S    &          &             \\     
                     & VI-split & VI-merge & Rand Error & VI-Split & VI-merge & Rand Error  \\
        \midrule
        Timepoint 45 & 0.3596   & 0.5918   & 0.1641     & 0.3740   &  0.5527  & 0.1517      \\
        Timepoint 49 & 0.4586   & 0.7116   & 0.2019     & 0.5153   &  0.5485  & 0.1873      \\
        \bottomrule
    \end{tabular}
    \caption{Comparison of Multciut and Lifted Multicut segmentation results for two time points taken from the light-sheet root primoridia data.}
    \label{tab_res_plant}
\end{table}

\section{Discussion} \label{sec_6}

We propose a general purpose strategy to leverage domain-specific knowledge for instance segmentation problems arising from EM image analysis.
This strategy makes use of a graph partitioning problem known as lifted multicut by expressing the domain knowledge in the long-range lifted edges.
We apply the proposed strategy to a diverse set of instance segmentation problems in light and electron microscopy and consistently show an improvement in segmentation accuracy.
For an application with ultrastructure based priors, we also observe that the lifted multicut based formulation yields higher quality results than the AMWC formulation of \citep{krasowski2017neuron}. We believe that this is due to joint exploitation of dense short-range and sparse long-range information. A complete joint solution, with both lifted edges and semantic labels, has recently been introduced in \citep{levinkov2017joint}. We look forward to exploring the potential of this objective for the neuron segmentation problem. 

Similar to the findings of \citep{kroeger2014asymmetric}, we demonstrated that prevention of merge errors is more efficient than their correction: the joint solution of LMC-S is more accurate than iterative LMC-SI. However, not all prior information can be incorporated directly into the original segmentation problem. For these priors we demonstrate how to construct an additional resolving step which can also significantly reduce the number of false merge errors. In the future we plan to further improve our segmentations by other sources of information: matches of the segmented objects to known cell types, manual skeletons or correlative light microscopy imaging.

\section{Acknowledgements}
We gratefully acknowledge the support of the Baden-Wuerttemberg Stiftung, and the contributions to this work made by Klaske J. Schippers and Nicole L. Schieber in the Electron Microscopy Facility of EMBL.

\clearpage
\bibliographystyle{frontiersinSCNS_ENG_HUMS}
\bibliography{bibliography}

\newpage
\section{Appendix}

\subsection{Overview of problem set-up}

\begin{table}[h]
    \center
    \begin{tabular}{l l l l}
        \toprule
                                   & Normal Edges                & Dense Lifted Edges                     & Sparse Lifted Edges \\
		\midrule
        Drosophila Neural Tissue   & Mean boundary evidence      & -                                      & \pbox{20 cm}{False merge \\ oracle predictions} \\ 
        Murine Neural Tissue       & RF based on edge features   & \pbox{20 cm}{RF based on region/ \\ clustering features} & \pbox{20 cm}{RF based on axon/ \\ dendrite attribution} \\ 
        Sponge Choanocytes         & Mean boundary evidence      & -                                      & \pbox{20 cm}{semantic segmentation of \\ small structures} \\
        Arabidopsis Roots          & Mean boundary evidence      & -                                      & \pbox{20 cm}{instance segmentation \\ of nuclei} \\
		\bottomrule
    \end{tabular}
    \caption{Overview of the three problem set-ups. RF abbreviates random forest.}
    \label{tab_res_overview}
\end{table}

\subsection{Hierarchical Lifted Multciut Solver} \label{lmc_solver}

\begin{algorithm}[h]
    \SetAlgoLined
    \KwData{graph $G$, edge weights $W_E$, lifted edges and weights $F$ and $W_F$, nLevels, blockShape}
    \KwResult{node partition $P$}
    $\hat G, \hat F, \hat W_E, \hat W_F$ = $G, F, W_E, W_F$\;
    \For{n in nLevels}{
        \nl blocks = getBlocks(blockShape)\;
        subPartitions = []\;
        \tcc{this for-loop can be parallelized}
        \For{block in blocks}{
            \nl $G_{sub}, W_E^{sub}$ = getSubproblem($\hat G$, $\hat W_E$, block)\;
            \nl $F_{sub}, W_F^{sub}$ = getLiftedEdges($G_{sub}$, $\hat F$, $\hat W_F$)\;
            \nl $P_{sub}$ = solveLiftedMulticut($G_{sub}, W_E^{sub}, F_{sub}, W_F^{sub}$)\;
            subPartitions.append($P_{sub}$)\;
        }
        \nl $\hat G, \hat F, \hat W_E, \hat W_F$ = reduceProblem($\hat G, \hat F, \hat W_E, \hat W_F$, subPartitions)\;
        blockShape *= 2\;
    }
    $P$ = solveLiftedMulticut($\hat G, \hat F, \hat W_E, \hat W_F$)\;
    $P$ = projectToInitialGraph($G, P$)\;
    \caption{Hierarchical lifted multicut algorithm based on the approximate multicut solver of \citep{pape2017solving}. 
             (1): getBlocks tiles the volume with blocks of blockShape.
             (2): getSubproblem extracts the sub-graph and weights of edges in this graph from the given block coordinates.
             (3): getLiftedEdges extracts the lifted edges that connect nodes which are \emph{both} part of the sub-graph as well as the corresponding weights.
             (4): solveLiftedMulticut solves the lifted multicut problem using one of the two approximate solvers \citep{beier2017multicut, keuper2015efficient}.
             (5): reduceProblem: reduces the graph by contracting nodes according to the sub-partition results. Also updates edge weights as well as lifted edges and their weights accordingly.}
    \label{algo_lmc}
\end{algorithm}

\begin{table}[h]
    \center
    \begin{tabular}{l l l}
    \toprule
                                              & Energy      & Time [s]  \\
    Greedy-Additive \citep{beier2016efficient} & -1585593.5  & 2.03     \\
    Kernighan-Lin \citep{keuper2015efficient}  & -1645876.7  & 174.69   \\
    Fusion-Moves \citep{beier2016efficient}    & -1645876.7  & 181.48   \\
    Hierarchical (Ours)                       & -1630274.3  & 3.29     \\
    \bottomrule
    \end{tabular}
    \caption{Evaluating our proposed hierarchical solver and other multicut solvers. In order to run this experiment, we have constructed a smaller lifted multicut problem from the Drosophila neural tissue dataset by cutting out a 1 $\times$ 10 $\times$ 10 micron block from its center, computing graph and local edge weights as described in \autoref{sec5}, introducing dense lifted edges within a graph neighborhood of 2 and setting their costs to the most repulsive edge cost along the weighted shorted path between the edge's terminal nodes. The problem at hand contained approximately 34,000 nodes, 244,000 normal edges and 2,384,000 lifted edges. The evaluation shows that the proposed solver yields energies comparable to Kernighan-Lin or Fusion-Moves, but its runtime is two orders of magnitude smaller and comparable to Greedy-Additive (which yields inferior energies). Kernighan-Lin was warm-started with the results of Greedy-Additive and Fusion-Moves with the results of Kernighan-Lin. Hierarchical has used Kernighan-Lin (warm-started with the solution of Greedy-Additive) for the sub-problems. While we only compare the solvers for a single problem size, we have observed very good scalability of our solver, which has solved much larger problems in \autoref{sec5}.}
    \label{tab_res_lmc}
\end{table}


\end{document}